\documentclass[acmtog]{acmart}
\AtBeginDocument{%
  }

\begin{document}

\title{Machine Learning (ML) library in Linux kernel}

\author{Viacheslav Dubeyko}
\email{Slava.Dubeyko@ibm.com}
\affiliation{%
  \institution{IBM}
  \city{San Jose}
  \state{California}
  \country{USA}
}

\renewcommand{\shortauthors}{Dubeyko et al.}

\begin{abstract}
  Linux kernel is a huge code base with enormous number of subsystems and possible configuration options that results in unmanageable complexity of elaborating an efficient configuration. Machine Learning (ML) is approach/area of learning from data, finding patterns, and making predictions without implementing algorithms by developers that can introduce a “self-evolving” capability in Linux kernel. However, introduction of ML approaches in Linux kernel is not easy way because there is no direct use of  floating-point operations (FPU) in kernel space and, potentially, ML models can be a reason of significant performance degradation in Linux kernel. Paper suggests the ML infrastructure architecture in Linux kernel that can solve the declared problem and introduce of employing ML models in kernel space. Suggested approach of kernel ML library has been implemented as Proof Of Concept (PoC) project with the goal to demonstrate feasibility of the suggestion and to design the interface of interaction the kernel-space ML model proxy and the ML model user-space thread.
\end{abstract}





\maketitle

\section{Introduction}
Nowadays, real life is a rapidly changing world that generates massive amount of heterogeneous digital data. Data accessing and processing workloads become more complex with unpredictable outcomes. Software implementation lifecycle always has time-consuming and resource-hungry nature that requires significant amount of time for analyzing problem, identifying an approach, implementing, and stabilizing the solution.  Generally speaking, we are running out of time and resources to deliver a required solution on time to manage the complexity and requirements of real-life workloads.

Linux kernel is a huge code base with enormous number of subsystems and possible configuration options. Finally, complexity and changing nature of modern workloads result in complexity of elaborating an efficient configuration and state of running Linux kernel. It is possible to say that Linux kernel requires to have a “self-evolving” capability.

Machine Learning (ML) is approach/area of learning from data, finding patterns, and making predictions without implementing algorithms by developers. The number of areas of ML applications is growing with every day. Generally speaking, ML can introduce a self-evolving and self-learning capability in Linux kernel.  There are already research works and industry efforts to employ ML approaches for configuration and optimization the Linux kernel. However, introduction of ML approaches in Linux kernel is not so simple and straightforward way. There are multiple problems and unanswered questions on this road. First of all, any ML model requires the floating-point operations (FPU) for running. But there is no direct use of FPUs in kernel space. Also, ML model requires training phase that can be a reason of significant performance degradation of Linux kernel. Even inference phase could be problematic from the performance point of view on kernel side. The using of ML approaches in Linux kernel is inevitable step. But, how can we use ML approaches in Linux kernel? Which infrastructure do we need to adopt ML models in Linux kernel?

\section{Related Works}
The number of areas of Machine Learning (ML) approaches application is growing with every day. There are already research works and industry efforts to employ ML approaches for configuration and optimization of Linux kernel. Shankar et al. \cite{Shankar2025} made comprehensive survey of potential applications of ML models in the area of Linux kernel optimization. This work highlights that integrating ML into Linux kernel holds great promise for enhancing system performance, resource management, and energy efficiency. Also, authors point out the challenges like difficulties  for developers to trust and understand the automated decisions being made, real-time ML inference often incurs high computational overhead,  data privacy is concern.

Several recent works have explored the eBPF technology for deploying ML models into Linux kernel from different perspectives. Gallego-Madrid et al. \cite{10.1007/978-981-97-4465-7_15} study  the opportunities of bringing  the extended Berkeley Packet Filter (eBPF) and Machine Learning (ML) technologies into Software Defined Networking (SDN) and Network Function Virtualization (NFV) field. Authors analyze the capability and challenges of  integration  of intelligent ML-powered programs into the Linux kernel for traffic processing, security analysis, etc. Chen et al. \cite{10.1145/3409963.3410492} explored the application of machine learning to the OS load balancing algorithm of a multiprocessor system. Authors demonstrated practical applications of ML for load balancing in the Linux kernel, utilizing eBPF for runtime system data collection through dynamic kernel tracing. The offline training in user space was employed  with online inference performed directly in the kernel to generate migration decisions. Sodhi et al. \cite{10.1145/3748355.3748363} argue that eBPF is the missing substrate for moderately complex in-kernel ML. Authors presented the eBPFML, a design that extends the eBPF instruction set with matrix-multiply helpers to facilitate efficient neural network inference. Wang et al. \cite{wang2024ebpfmeetsmachinelearning} presented O2C, a system for on-the-fly OS kernel compartmentalization that embeds machine learning models (specifically, a decision tree) into eBPF programs. Their work demonstrates that ML models can learn invariants in various kernel object types, with runtime inference used to extract and classify heap object contents. Using of ML models in the Linux kernel has shown particular promise for real-time security applications. Brodzik et al. \cite{brodzik2024ransomwaredetectionusingmachine} investigated the integration of two ML algorithms, decision tree and multilayer perceptron, in eBPF with the objective of enhancing early ransomware detection in Linux environments.  Rather than sending event data to user space for analysis, they implement the entire threat detection process directly in eBPF.

Zhang et al. \cite{10.1145/3544216.3544229} presented LiteFlow, a hybrid solution to build high-performance adaptive neural networks (NN) for kernel datapath.  LiteFlow decouples the control path of adaptive NNs into: (1) a kernel-space fast path for efficient model inference, and (2) a userspace slow path for effective model tuning.  LiteFlow has been applied to congestion control, flow scheduling, and load balancing, demonstrating that adaptive neural networks can operate effectively in kernel datapath.

\section{ML Library Architecture}
What is the goal of using ML models in Linux kernel? The main goal is to employ ML models for elaboration of a logic of particular Linux kernel subsystem based on processing data or/and an efficient subsystem configuration based on internal state of subsystem. As a result, it needs: (1) collect data for training, (2) execute ML model training phase, (3) test trained ML model, (4) use ML model for executing the inference phase. The ML model inference can be used for recommendation of Linux kernel subsystem configuration or/and for injecting a synthesized subsystem logic into kernel space (for example, eBPF logic).

Usually, ML model requires floating point operations (FPU) and Python is widely used for implementing and running ML models. As a result, ML model cannot be run directly in the Linux kernel. Potentially, ML model can be modified to run in kernel space without  floating point operations. However, it makes ML model inaccurate and it could dramatically decrease the efficiency of using ML approaches in Linux kernel. From another point of view, running ML models in kernel space can be a reason of Linux kernel performance degradation (especially, during training phase). Moreover, the continuous learning approach could be a beneficial model for the case of dynamic nature of modern and future workloads. Generally speaking, currently, running ML models directly in Linux kernel is very problematic and not efficient way. Running ML models in user-space is more promising direction because they still can be implemented in Python and it can use the  floating point operations in user-space. ML model in user-space can be considered like a Linux kernel extension. As a result, kernel subsystem will be able to interact with user-space process/thread by means of sysfs, FUSE, or character device, for example.

How ML infrastructure can be designed in Linux kernel? It needs to introduce in Linux kernel a special ML library that can implement a generalized interface of interaction between ML model’s thread in user-space and kernel subsystem that can be extended or specialized by a particular kernel subsystem. Likewise interface requires to have the means: (1) create, initialize, destroy ML model proxy in kernel subsystem, (2) start, stop ML model proxy, (3) get, preprocess, publish data sets from kernel space, (4) receive, preprocess, apply ML model recommendation(s) from user-space, (5) execute synthesized logic or recommendations in kernel-space, (6) estimate efficiency of synthesized logic or recommendations, (7) execute error back-propagation with the goal of correction ML model on user-space side.

The create and initialize logic can be executed by kernel subsystem during module load or Linux kernel start (oppositely, module unload or kernel shutdown will execute destroy of ML model proxy logic). ML model thread in user-space will be capable to re-initialize and to execute the start/stop logic of  ML model proxy on kernel side. First of all, ML model needs to be trained by data from kernel space. The data can be requested by ML model from user-space or data can be published by ML model proxy from kernel-space. The sysfs interface can be used to orchestrate this interaction. As a result, ML model in user-space should be capable to extract data set(s) from kernel space through sysfs, FUSE or character device. Extracted data can be stored in persistent storage and, finally, ML model can be trained in user-space by accessing these data.

The continuous learning model can be adopted during training phase. It implies that kernel subsystem can receive ML model recommendations even during training phase. ML model proxy on kernel side can estimate the current kernel subsystem state, tries to apply the ML model recommendations, and estimate the efficiency of applied recommendations. Generally speaking, ML model proxy on kernel side can consider several modes of interaction with ML model recommendations: (1) emergency mode, (2) learning mode, (3) collaboration mode, (4) recommendation mode. The emergency mode is the mode when kernel subsystem is in critical state and it is required to work as efficient as possible without capability of involving the ML model recommendations (for example, ML model recommendations are completely inadequate or load is very high). The learning mode implies that kernel subsystem can try to apply the ML model recommendations for some operations with the goal of estimation the maturity of ML model. Also, ML model proxy can degrade the mode to learning state if ML model recommendations becomes inefficient. The collaboration mode has the goal of using ML recommendations in 50\% of operations with the goal of achieving mature state of ML model. And, finally,  ML model proxy can convert kernel subsystem in recommendation mode if ML model is mature enough and efficiency of applying the ML recommendations is higher than using human-made algorithms. The back-propagation approach can be used to correct the ML model by means of sharing feedback of efficiency estimation from kernel to user-space side.

\section{Conclusion and Contribution}
Suggested approach of kernel ML library has been implemented as Proof Of Concept (PoC) project \cite{ml_lib_source_code, ml_lib_kernel} with the goal to demonstrate feasibility of the suggestion and to design the interface of interaction the kernel-space ML model proxy and the ML model user-space thread. The repository \cite{ml_lib_source_code, ml_lib_kernel} contains examples of generalized and specialized ML model proxy. As the future work, we plan to implement ML-based GC subsystem for several kernel-space file systems (F2FS, NILFS2, SSDFS) and ML-based DAMON extension with the goal to check the approach in real-life applications and to show the capability to achieve better efficiency by employing a “self-learning” ML model(s) in Linux kernel.

\bibliographystyle{ACM-Reference-Format}
\bibliography{sample-base}

@String{Computing = "Computing" }

@String{Computer = "{IEEE} Computer" }

@String{Springer = "Springer-Verlag" }

@online{ml_lib_source_code,
  author =    {Viacheslav Dubeyko},
  title =  {ML library source code},
  year = 2026,
  url =    {https://github.com/kernel-ml-lib/ml-lib},
  lastaccessed = {January 16, 2026}
  }

@online{ml_lib_kernel,
  author =    {Viacheslav Dubeyko},
  title =  {Linux kernel with ML library},
  year = 2026,
  url =    {https://github.com/kernel-ml-lib/ml-lib-linux},
  lastaccessed = {January 16, 2026}
  }

@article{Shankar2025,
author = {Shankar, Vasuki},
year = {2025},
month = {03},
pages = {56-64},
title = {Machine Learning for Linux Kernel Optimization: Current Trends and Future Directions},
volume = {13},
journal = {International Journal of Computer Sciences and Engineering},
doi = {10.26438/ijcse/v13i3.5664}
}

@InProceedings{10.1007/978-981-97-4465-7_15,
author="Gallego-Madrid, Jorge
and Bru-Santa, Irene
and Sanchez-Iborra, Ramon
and Skarmeta, Antonio",
editor="You, Ilsun
and Chora{\'{s}}, Micha{\l}
and Shin, Seonghan
and Kim, Hwankuk
and Astillo, Philip Virgil",
title="Integrating Machine Learning Models into the Linux Kernel: Opportunities and Challenges",
booktitle="Mobile Internet Security",
year="2024",
publisher="Springer Nature Singapore",
address="Singapore",
pages="209--219",
isbn="978-981-97-4465-7"
}

@inproceedings{10.1145/3409963.3410492,
author = {Chen, Jingde and Banerjee, Subho S. and Kalbarczyk, Zbigniew T. and Iyer, Ravishankar K.},
title = {Machine learning for load balancing in the Linux kernel},
year = {2020},
isbn = {9781450380690},
publisher = {Association for Computing Machinery},
address = {New York, NY, USA},
url = {https://doi.org/10.1145/3409963.3410492},
doi = {10.1145/3409963.3410492},
booktitle = {Proceedings of the 11th ACM SIGOPS Asia-Pacific Workshop on Systems},
pages = {67–74},
numpages = {8},
location = {Tsukuba, Japan},
series = {APSys '20}
}

@inproceedings{10.1145/3748355.3748363,
author = {Sodhi, Prabhpreet Singh and Liargkovas, Georgios and Kaffes, Kostis},
title = {Empowering machine-learning assisted kernel decisions with eBPFML},
year = {2025},
isbn = {9798400720840},
publisher = {Association for Computing Machinery},
address = {New York, NY, USA},
url = {https://doi.org/10.1145/3748355.3748363},
doi = {10.1145/3748355.3748363},
booktitle = {Proceedings of the 3rd Workshop on EBPF and Kernel Extensions},
pages = {28–30},
numpages = {3},
location = {Coimbra, Portugal},
series = {eBPF '25}
}

@misc{wang2024ebpfmeetsmachinelearning,
      title={When eBPF Meets Machine Learning: On-the-fly OS Kernel Compartmentalization}, 
      author={Zicheng Wang and Tiejin Chen and Qinrun Dai and Yueqi Chen and Hua Wei and Qingkai Zeng},
      year={2024},
      eprint={2401.05641},
      archivePrefix={arXiv},
      primaryClass={cs.OS},
      url={https://arxiv.org/abs/2401.05641}, 
}

@misc{brodzik2024ransomwaredetectionusingmachine,
      title={Ransomware Detection Using Machine Learning in the Linux Kernel}, 
      author={Adrian Brodzik and Tomasz Malec-Kruszyński and Wojciech Niewolski and Mikołaj Tkaczyk and Krzysztof Bocianiak and Sok-Yen Loui},
      year={2024},
      eprint={2409.06452},
      archivePrefix={arXiv},
      primaryClass={cs.CR},
      url={https://arxiv.org/abs/2409.06452}, 
}

@inproceedings{10.1145/3544216.3544229,
author = {Zhang, Junxue and Zeng, Chaoliang and Zhang, Hong and Hu, Shuihai and Chen, Kai},
title = {LiteFlow: towards high-performance adaptive neural networks for kernel datapath},
year = {2022},
isbn = {9781450394208},
publisher = {Association for Computing Machinery},
address = {New York, NY, USA},
url = {https://doi.org/10.1145/3544216.3544229},
doi = {10.1145/3544216.3544229},
booktitle = {Proceedings of the ACM SIGCOMM 2022 Conference},
pages = {414–427},
numpages = {14},
location = {Amsterdam, Netherlands},
series = {SIGCOMM '22}
}

%\appendix

\end{document}